%% file: neurips_2025.tex
\documentclass{article}
\usepackage{float}

\usepackage[preprint]{neurips_2025}

\usepackage[utf8]{inputenc} 
\usepackage[T1]{fontenc}    
\usepackage{hyperref}       
\usepackage{url}            
\usepackage{booktabs}       
\usepackage{multirow}       
\usepackage{amsfonts}       
\usepackage{nicefrac}       
\usepackage{microtype}      
\usepackage{xcolor}         
\usepackage{wrapfig}
\usepackage{algorithm}
\usepackage{algorithmic}
\usepackage{graphicx}
\usepackage{mathrsfs}
\usepackage{amsmath}
\usepackage{xspace}
\usepackage{pifont}
\usepackage{array}
\usepackage{subcaption} 
\definecolor{modblue}{RGB}{100,150,220}    
\definecolor{modorange}{RGB}{255,180,100}  
\newcommand{\cmark}{\ding{51}}%
\newcommand{\xmark}{\ding{55}}%
\newcommand{\method}{\text{DualParal}\xspace}

\title{Minute-Long Videos with Dual Parallelisms}

\author{
    \vspace{-2em}
    \\
    \href{https://dualparal-project.github.io/dualparal.github.io/}{
    https://dualparal-project.github.io/dualparal.github.io/
    }
    \\
    Zeqing Wang$^{12}$
    \quad 
    Bowen Zheng$^{13}$
    \quad 
    Xingyi Yang$^1$
    \quad
    Zhenxiong Tan$^1$
    \\
    Yuecong Xu$^1$
    \quad 
    Xinchao Wang\thanks{Corresponding Author}~~$^1$
    \\
    $^1$National University of Singapore
    \\
    $^2$Xidian University
    \quad
    $^3$Huazhong University of Science and Technology
    \\
    \texttt{zeqing.wang@stu.xidian.edu.cn \quad xinchao@nus.edu.sg} \\
}

\begin{document}

\maketitle

\input{sections/1_abstract}   
\input{sections/2_introduction}
\input{sections/4_preliminaries}

\input{sections/5_method}

\input{sections/6_experiments}

\input{sections/7_conclusion}

{
    \small
    \bibliographystyle{plain}
    \bibliography{main}
}



\appendix
\input{sections/8_appendix}

\end{document}

%% file: sections/1_abstract.tex
\begin{abstract}

Diffusion Transformer (DiT)-based video diffusion models generate high-quality videos at scale but incur prohibitive processing latency and memory costs for long videos.
To address this, we propose a novel distributed inference strategy, termed \textbf{\method}. The core idea is that, instead of generating an entire video on a single GPU, we parallelize both temporal frames and model layers across GPUs. However, a naive implementation of this division faces a key limitation: since diffusion models require synchronized noise levels across frames, 
this implementation leads to the serialization of original parallelisms. 
We leverage a \textit{block-wise denoising} scheme to handle this. Namely, we process a sequence of frame blocks through the pipeline with progressively decreasing noise levels. Each GPU handles a specific block and layer subset while passing previous results to the next GPU, enabling asynchronous computation and communication.
To further optimize performance, we incorporate two key enhancements. Firstly, a feature cache is implemented on each GPU to store and reuse features from the prior block as context, minimizing inter-GPU communication and redundant computation. Secondly, we employ a coordinated noise initialization strategy, ensuring globally consistent temporal dynamics by sharing initial noise patterns across GPUs without extra resource costs. 
Together, these enable fast, artifact-free, and infinitely long video generation. Applied to the latest diffusion transformer video generator, our method efficiently produces 1,025-frame videos with up to 6.54$\times$ lower latency and 1.48$\times$ lower memory cost on 8$\times$RTX 4090 GPUs.

\end{abstract}

%% file: sections/2_introduction.tex
\section{Introduction}\label{sec:intro}

Diffusion Transformer (DiT)~\cite{DiT} has significantly improved the scalability of video diffusion models~\cite{hunyuan, Wan, opensora2}, enabling more realistic and higher-resolution video generation. 
Despite its benefits, large-scale DiT models suffer from their inherent computational inefficiency. This directly results in extended processing durations and memory demands, especially for real-time deployment~\cite{xDit}.


Notably, this inefficiency is further exacerbated when generating long videos. Intuitively, longer videos increase the input sequence length. This has severe implications for latency: the attention mechanism, core to DiT~\cite{xDit, Pipefusion}, exhibits time complexity that scales quadratically with sequence length. Concurrently, memory consumption escalates substantially due to the combination of a large number of model parameters and the extended video sequences. Therefore, enabling DiT-based models to efficiently generate high-quality long videos remains a formidable and pressing challenge.
Recently, parallelization has emerged as a promising solution for efficient long video generation. It uses multiple devices to produce video jointly, which scales memory and boosts processing speed.
Among existing strategies, \textit{sequence parallelism} reduces latency by synchronously processing split hidden~\cite{Ulysses, RingAttention} or input~\cite{infinity, FIFO} sequences using a full model replica on each device. However, they incurs high memory overhead due to the entire model on every device~\cite{infinity, FIFO}. In contrast, \textit{pipeline parallelism}~\cite{Pipefusion} mitigates memory usage by partitioning the model across devices as a device pipeline~\cite{Gpipe, TeraPipe, Megatron}. Therefore, an ideal solution would combine the sequence parallelism with pipeline parallelism to maximize speed and minimize memory usage.

However, naively combining sequence and pipeline parallelism is fundamentally conflicting. The core issue stems from the inherent \emph{synchronization property} of video diffusion models: all input tokens must pass through an entire layer together before any can move on. In pipeline parallelism, this means the full input must finish processing on one device (e.g., Device 1) before passing to the next (e.g., Device 2). This requirement directly contradicts sequence parallelism, which splits the input across devices.
As a result, all distributed parts must be gathered back onto a single device for serialized processing on specific model layers. Only then can all parts enter the next pipeline stage, i.e. next device. This repeated gathering serializes computation and negates the benefits of sequence parallelism, reintroducing a serial bottleneck and significant communication overhead.
To address this conflict, we propose a novel distributed inference strategy, termed \textbf{\method}. 
At a high level, \method divides both the video sequence and model into chunks and applies parallel processing across both. 

As discussed above, a naive combination presents challenges. Inspired by recent work on interpolating diffusion and autoregressive models~\cite{arriola2025block,magi1}, we make this feasible by implementing a \emph{block-wise denoising} scheme for video diffusion models. By the word \emph{block-wise}, we refer to a strategy where, instead of denoising all frames at a uniform noise level, we divide the video into non-overlapping temporal blocks. 
Each block is assigned a different noise level according to its position in the video: blocks closer to the end have higher noise levels, while earlier blocks receive lower noise levels.
During each inference step, the model processes all blocks 
asynchronously,
incrementally reducing their respective noise levels. Crucially, because noise levels do not need to be synchronized across all frames, block-wise denoising resolves the inherent conflict between the two parallelism strategies.

Accordingly, we tailor this inference scheme for multiple devices in conjunction with our \method. Specifically, we organize video sequence blocks in a first-in-first-out (FIFO) queue~\cite{FIFO, rolling}, where noise levels decrease from tail to head. In each diffusion step, a new noisy block is appended to the tail, while a clean block is removed from the head. These video blocks are then processed in reverse order, tail to head, through a device pipeline. 
In this setup, each device handles a specific video block and a model chunk, with denoised outputs passed asynchronously between GPUs. This distributed architecture enhances memory efficiency by distributing the model and achieves near-zero idle time through asynchronous block processing and communication.

Even more compelling, \method leverages its FIFO queue to enable long video generation. New blocks can be continuously appended to the queue, allowing for producing arbitrarily long videos. Because the number of frames within each block remains fixed, this approach again avoids quadratic increase in processing latency and high memory costs associated with extended video sequences.

To further optimize parallel efficiency and maintain video quality, we introduce two key enhancements for \method. 
\textit{Firstly}, to ensure coherence between adjacent blocks, each block is concatenated with parts of previous and subsequent blocks before processed in the device pipeline, resulting extra resource costs. To reduce this, \method employs a feature cache on each GPU that stores and reuses Key-Value (KV) features from the previous block without explicitly concatenating it. This reduce inter-GPU communication and redundant computation in components like Cross-Attention~\cite{Attention, Wan} and Feed-Forward Networks (FFN)~\cite{Attention} in the latest video diffusion model Wan2.1~\cite{Wan}.
\textit{Secondly}, to maintain global consistency across blocks without extra resource costs for global information, new block is initialized with a coordinated noise space, avoiding performance degradation caused by repetitive noise, all without extra cost.
Together, these enable fast, artifact-free, and infinite-length video generation. 

In summary, our contributions are summarized as follows: (1) We design an efficient distributed inference strategy by parallelizing both the video sequence and model layers, operating under a \textit{block-wise denoising} scheme, to minimize idle time and optimize both computation and memory usage. (2) We employ feature cache and coordinated noise initialization strategies to optimize parallel efficiency while preserving video quality. (3) Experiments show that \method achieves up to a 6.54$\times$ reduction in latency and a 1.48$\times$ reduction in memory cost compared to state-of-the-art distributed methods when generating 1,025-frame videos with 8$\times$RTX 4090 GPUs.

%% file: sections/4_preliminaries.tex
\section{Preliminaries}\label{sec:preliminaries}

\textbf{Diffusion models in video generation}

Video generation using diffusion models~\cite{diffusion, DiT, Wan, hunyuan, cogvideo} involves progressively denoising frame latent representation $x_t$, where $t$ denotes the noise level and ranges from $T$ (the most noisy state) to $0$ (the cleanest). Here, $T$ also represents the total number of denoising steps. The process starts with a complete noisy latent $x_T$, and through each denoising step, $x_t$ is updated to a clearer $x_{t-1}$. This continues until $x_T$ is denoised to $x_0$, which is then decoded to generate the final video. The key operation in updating $x_t$ to $x_{t-1}$ involves computing the noisy prediction $\epsilon_t = \mathcal{E}_{\theta}(x_t)$, where $\mathcal{E}_{\theta}$ represents the diffusion model. Subsequently, $x_{t-1}$ is derived using $x_{t-1} = S(\epsilon_t, x_t, t)$, where $S$ is the updating scheduler of the corresponding video diffusion model.

Specifically, the noisy frame latent is defined as $x_t \in \mathbb{R}^{F \times H \times W \times C}$, where $F$ represents the number of frames, $H$ and $W$ are the height and width of each frame latent, respectively, and $C$ is the number of channels. It is important to note that after passing $x_0$ through the decoder to generate the final video $X$, the dimensions of $X$ differ from those of $x_0$ due to the upsampling process in the decoder~\cite{Wan, hunyuan, opensora2}. For simplicity, we use the dimensions of $x$ to represent the video.

\textbf{Parallelisms for DiT-based video diffusion models}

Pipeline parallelism \cite{Gpipe, TeraPipe, zerobuuble} typically involves evenly splitting the entire neural network across $N$ devices, with each device responsible for a consecutive subset of the model, denoted as $\mathcal{E}_{\theta} = [\mathcal{E}_{\theta_1}, \mathcal{E}_{\theta_2}, \dots, \mathcal{E}_{\theta_N}]$. Since DiT-based video diffusion models~\cite{DiT, Wan} are generally composed of multiple similar DiT blocks, we define $L$ as the total number of DiT blocks, with each device handling consecutive $\frac{L}{N}$ DiT blocks. Therefore, denoising $x_t$ is represented as:
\begin{equation}
\epsilon_t = \mathcal{E}_{\theta_N}(\mathcal{E}_{\theta_{N-1}}( \dots (\mathcal{E}_{\theta_1}(x_t)) \dots )) = \mathcal{E}_{\theta_N}( \dots (\mathcal{E}_{\theta_j}(\epsilon_t^{j-1})) \dots ),
\label{eq:denoising_parallel}
\end{equation}
where $\epsilon_t^{j-1}\in\mathbb{R}^{p \times h}$ denotes the noisy prediction from the previous $(j-1)^{th}$ device. Here, $p$ represents the sequence length and $h$ denotes the hidden size. Specifically, $p = F' \times H' \times W'$, where $F'$, $H'$, and $W'$ are the downsampled dimensions of $F$, $H$, and $W$, respectively. For the Wan2.1 model~\cite{Wan} used as the base in this paper, $F' = F$. Therefore, we define $p = F \times H' \times W'$.

Sequence parallelism divides the input $x_t \in \mathbb{R}^{F \times H \times W \times C}$ into non-overlapping blocks, each denoted as $B_t \in \mathbb{R}^{Num_B \times H \times W \times C}$, where $Num_B$ is the number of frames per block. To enhance temporal coherence across adjacent blocks, several methods~\cite{FIFO, infinity} concatenate previous, subsequent, or global context frames with the current block during denoising, resulting in the extended block $B_t' \in \mathbb{R}^{(Num_B + Num_C) \times H \times W \times C}$, where $Num_C$ denotes the number of concatenated context frames.

%% file: sections/5_method.tex
\section{\method}\label{sec:method}
\begin{figure}[t]
    \centering
    \includegraphics[page=1, scale=0.55]{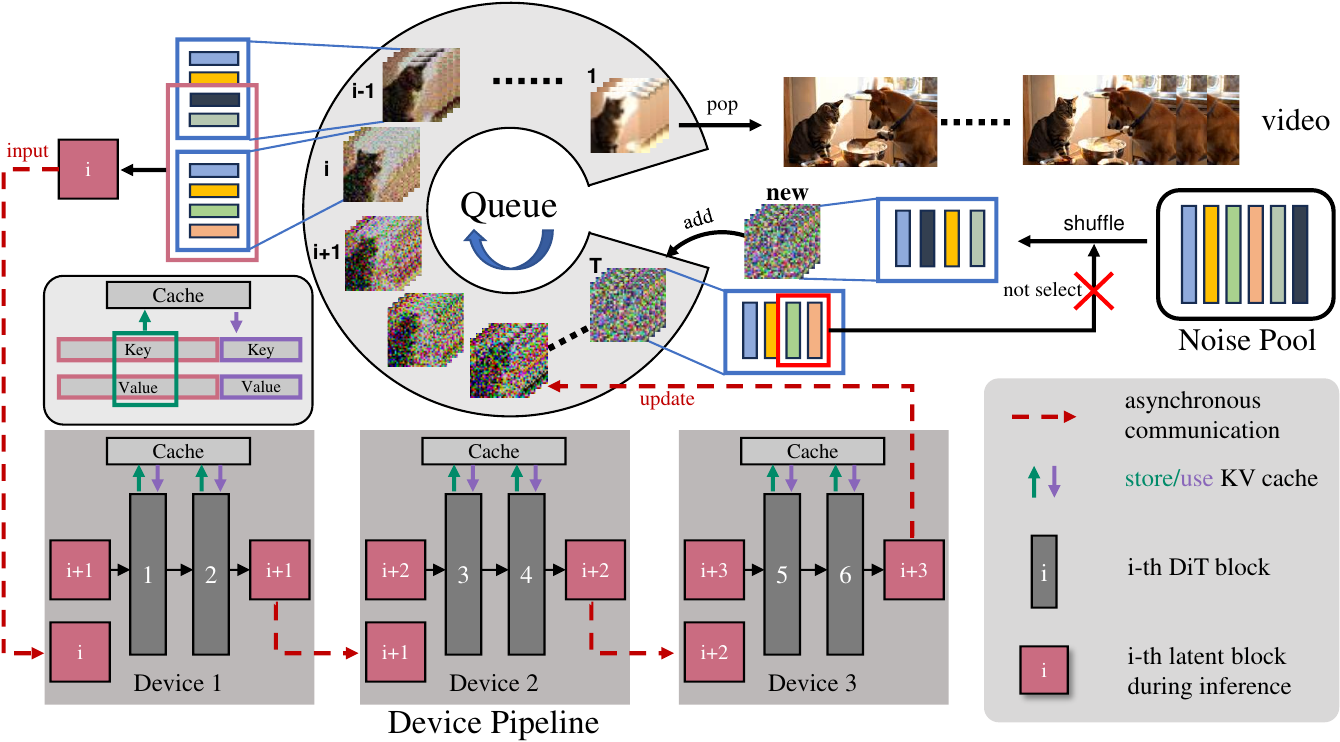} 
    \caption{Overview of \method: 
    \method partitions video frames into sequential blocks organized in a queue with noise levels increasing from tail to head, and distributes model layers across devices via a device pipeline.
    By feeding blocks into the pipeline in a reverse order (from tail to head), this \textit{block-wise denoising} scheme significantly improves efficiency. 
    To further improve performance, \method reuses Key-Value (KV) features from the previous block, requiring only the subsequent block to be concatenated. To preserve global consistency, each new block is initialized from a shared noise pool by shuffling noises, excluding the last $\frac{Num_c}{2}$ latents of the last block in queue.
    }
    \label{fig:DualPipe}
    \vspace{-1.2em}
\end{figure}

At a high level, \method introduces dual parallelisms over both the video sequence and model layers while leveraging a \textit{block-wise denoising} scheme to achieve computational and memory efficiency. An overview is provided in \autoref{fig:DualPipe}, with architectural details discussed in Section~\ref{sec:architecture}. To further improve efficiency, in Section~\ref{sec:feature_cache}, we design a feature cache that reuses KV features from the previous block, reducing inter-device communication and redundant computation. Additionally, in Section~\ref{sec:noise}, a coordinated noise initialization strategy is adopted to ensure global consistency without additional resource overhead. 
Lastly, for better illustration the efficiency of \method, we provide a theoretical analysis of parallel performance in Section~\ref{sec:Quantitative}. 
\subsection{Parallel architecture}\label{sec:architecture}
Naively combining sequence and pipeline parallelism introduces an inherent conflict: synchronizing noise levels across frames requires all split sequences to be gathered and processed on a single device before proceeding to the next device in the device pipeline. This conflict degrades both parallelism strategies into serialized processing.
As a result, such degradation leads to high device idle time in standard pipeline parallelism and breaks parallel execution in sequence parallelism. Moreover, it introduces significant communication overhead due to the repeated gathering of split sequences.

To address these issues, \method adopts dual parallelisms under a \textit{block-wise denoising} mechanism. Namely, \method simultaneously process block-wise frames with asynchronous noise levels across different model chunks. Since noise levels do not need to be synchronized across frames at each model segment, \method effectively resolves the conflict between sequence and pipeline parallelism.

Specifically, as illustrated in Figure~\ref{fig:DualPipe}, \method comprises two key components: queue and device pipeline. In device pipeline, DiT blocks from the video diffusion model are evenly distributed across multiple GPUs. 
Within the queue, each element is a block of $Num_B$ frame latents sharing the same noise level, formally denoted as $B_i = [x_i, x_i, \dots, x_i]$, where $x_i \in \mathbb{R}^{1 \times H \times W \times C}$ represents a single frame latent at the $i^\text{th}$ noise level.
Additionally, the queue is organized in a first-in-first-out manner~\cite{FIFO, rolling}, with blocks arranged from tail to head in progressively decreasing noise levels, ranging from 1 to $T$. Formally, queue is described as $Q=[B_1, B_2, \dots, B_T]$.
During inference, blocks in the queue are continuously fed into the device pipeline in reverse order, from tail to head. After each diffusion step, all blocks in the queue shift forward by one position, i.e., $Q = [B_0, B_1, \dots, B_{T-1}]$. A new noisy block $B_T$ is then appended to the tail, while the clean block $B_0$ is removed from the head and passed to the decoder for final video reconstruction.
With this implementation, each device handles a specific video block and a corresponding model segment, while denoised outputs are passed asynchronously between GPUs. This \textit{block-wise denoising} scheme effectively resolves the serialization degradation caused by naively combining sequence and pipeline parallelism, thereby enabling true parallelization across both temporal frames and model layers.

\subsection{Feature cache}\label{sec:feature_cache}
Since whole video frames are divided into non-overlapping blocks, we concatenate previous and subsequent blocks with a total of $Num_C$ frame latents to maintain temporal coherence, resulting in the denoising of an extended block $B_i' = [B_{i-1}, B_{i}, B_{i+1}]$. For simplicity, we assume that all frame latents from both adjacent blocks are included, i.e., $Num_C = 2Num_B$. Note that $B_{i-1}$ denotes the subsequent block, while $B_{i+1}$ refers to the previous block in the reversed inference order. 
Therefore, denoising block $B_i'$ is formally described as:
\begin{equation}
\epsilon_i= \mathcal{E}_{\theta_N}(\mathcal{E}_{\theta_{N-1}}( \dots (\mathcal{E}_{\theta_1}(B_i')) \dots )) = \mathcal{E}_{\theta_N}( \dots (\mathcal{E}_{\theta_j}(\epsilon_i^{j-1})) \dots ),
\end{equation}
where each intermediate output $\epsilon_i^{j-1}$ is transmitted from $(j-1)^{\text{th}}$ to $j^{\text{th}}$ device using asynchronous peer-to-peer (P2P) communication, allowing communication and computation to overlap effectively.

However, this implementation introduces extra communication and computation overhead due to the concatenated parts. To mitigate this, we exploit a unique feature of \method and propose a feature cache technique.
Specifically, since block $B_i'=[B_{i-1}, B_i, B_{i+1}]$ is denoised after the previous block $B_{i+1}'=[B_{i}, B_{i+1}, B_{i+2}]$, $B_{i+1}$ has already been processed during the denoising of $B_{i+1}'$. Leveraging this feature, we cache the KV features from the Self-Attention module of $B_{i+1}$ during denoising $B_{i+1}'$ and reuse them when denoising $B_i'$. Consequently, the input block for denoising is reduced to $B_i' = [B_{i-1}, B_i]$, decreasing communication overhead between adjacent devices.

Moreover, for $B_i'$, adjacent blocks $B_{i-1}$ and $B_{i+1}$ assist in denoising $B_i$. Among all model components, only those that require interaction across frames—such as the Self-Attention module in the Wan2.1 model~\cite{Wan}—contribute meaningfully in this context. Therefore, we restrict the feature caching technique to the Self-Attention module while skipping components like Cross-Attention and FFN, which do not benefit from inter-frame information. This selective application effectively eliminates redundant computations.

\subsection{Coordinated noise initialization}\label{sec:noise}
Although \method concatenates previous and subsequent blocks to smooth transitions, global consistency remains a challenge. A simple solution—concatenating more global information—incurs high communication, computation, and memory costs. To avoid these, reusing noisy latents from the same noise space~\cite{freenoise, videomerge} offers a promising alternative. This section analyzes different initialization methods to determine the best strategy for ensuring global consistency, specifically for DiT-based video diffusion models. There are two key observations: \textbf{1) Using \textit{complete} noise space maintains favorable global consistency. 2) Latents with the repetitive noise during the \textit{whole} denoising process cause significant performance degradation in the DiT-based video diffusion model.}

\begin{figure}[h]
    \centering
    \vspace{-0.5em}
    \includegraphics[page=1, width=\textwidth]{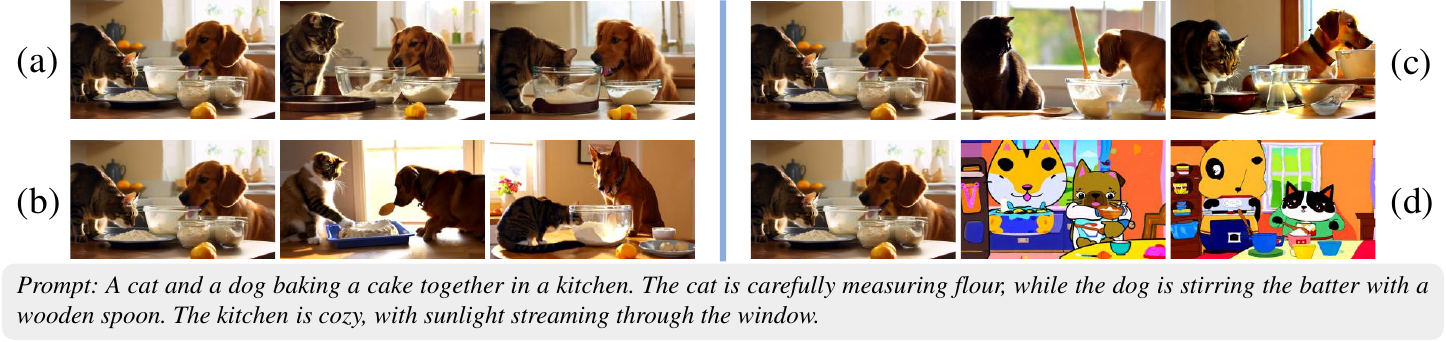}
    \caption{Examples of four different noise initializations for Wan2.1 model~\cite{Wan}: (a) uses the complete noise space, (b) uses a subset of the noise space, (c) adds new noise to the original space, and (d) uses the complete noise space with the repetitive noise. The first image shows the standard video generated from the reference noise space, followed by two different orders of noise initialization.}
    \label{fig:Noise}
    \vspace{-0.5em}
\end{figure}

The first observation, illustrated in Figure~\ref{fig:Noise}, shows that using the complete noise space (a) yields better global consistency compared to using a subset of noise (b) or adding new noise (c) to the original space. Based on this, we initialize blocks in \method using the complete noise space, with varying initialization orders. However, as shown in (d) of Figure~\ref{fig:Noise}, directly denoising using the same noise in Wan2.1, a DiT-based video diffusion model, leads to a significant performance degradation. This second observation arises from repetitive noises when concatenating the subsequent block in \method during \textit{whole} denoising process. In contrast, the previous block only affects the Self-Attention module in Wan2.1, without causing performance degradation.
To resolve this problem while preserving the complete noise space's advantages, we propose a novel initialization strategy. Specifically, as shown in \autoref{fig:DualPipe}, when initializing a new block, we select noise from a pool that has not been used in the last $\frac{Num_C}{2}$ latents of the final block $B_T$ in the queue (e.g., $Num_C = 4$ in \autoref{fig:DualPipe}). These selected noises are then shuffled and used to initialize the new block. Note that the first block uses the complete noise pool and contains $\frac{Num_C}{2} + Num_B$ frames. This strategy ensures that the same noise isn't reused in concatenated blocks during the whole denoising process, while still utilizing the complete noise pool throughout the process.

\subsection{Quantitative analysis of efficiency}\label{sec:Quantitative}
This section provide quantitative analysis of parallel performance of \method in terms of bubble ratio, communication overhead, and memory cost. 

\begin{figure}[h]
    \centering
    \vspace{-0.6em}
    \includegraphics[page=1, width=\textwidth]{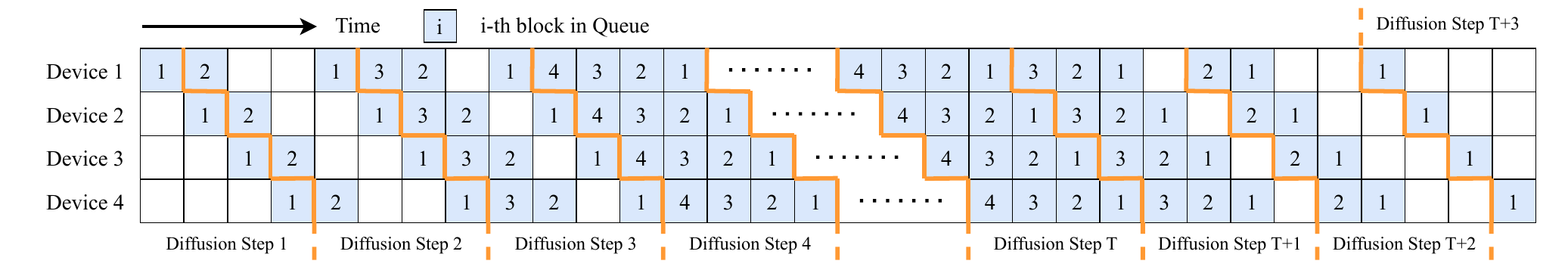}
    \caption{Pipeline schedule of \method with $N=4$, $T=50$, and $Block_{num}=4$. Blocks are denoised in reverse order, from tail to head in the queue. After diffusion step $T$, the first clean block is popped from the queue, and all remaining blocks shift forward by one position, decrementing their indices accordingly.}
    \label{fig:Bubble}
    \vspace{-0.5em}
\end{figure}

For bubble ratio~\cite{Gpipe}, it evaluate the ratio of idle time in each device. 
We compute it for \method under the reverse (tail-to-head) denoising order. Additionally, we assume $N \leq Block_{num}$, where $Block_{num}$ denotes the total number of blocks during long video generation. This assumption is reasonable, as it can be easily satisfied in practice for long videos, especially for minute-long videos. Therefore, the bubble ratio is formally expressed as:
\begin{equation}
    Bubble = \frac{Bubble\:Size}{Bubble\:Size + NonBubble\:Size} = \frac{N^2 - N-1}{N^2-N-1 + T \times Block_{num}}.
\label{eq:bubble}
\end{equation}
The detailed proof of \autoref{eq:bubble} is provided in Appendix~\ref{appendix:analysis_of_different_denoising_order}. To intuitively illustrate the bubble ratio, \autoref{fig:Bubble} presents an example of the pipeline scheduling in \method, exhibiting an approximate bubble ratio of $5.2\%$. Moreover, as $Block_{num}$ increases, the bubble ratio approaches $0\%$, indicating minimal device idle time in the pipeline during long video generation.

Idle time occurs during brief warm-up and cool-down phases, when the current number of blocks in the queue is smaller than the number of devices $N$—for example, before diffusion step 4 and after diffusion step $T$ in \autoref{fig:Bubble}. During these phases, some idle time and synchronization overhead may arise due to non-overlapping communication and computation. However, these periods are relatively short and thus contribute negligible overhead in the context of long video generation. Therefore, \method achieve high utilization of multiple GPUs. 
Further analysis of the bubble ratio—including detailed proofs and discussions under different denoising conditions (e.g., different denoising order, and the case where $N > Block_{num}$)—is provided in Appendix~\ref{appendix:analysis_of_different_denoising_order}.



\begin{table}[h]
\renewcommand{\arraystretch}{0.45}
\vspace{-1.1em}
\caption{Comparison of parallel methods for the DiT-based video diffusion model at a single diffusion step. `Overlap' refers to the degree of overlap between communication and computation. $W$ is total memory cost of the model, while $KV$ represents the memory cost for a single frame input.}
\label{tab:parallel_schemes}
\centering
\resizebox{\linewidth}{!}{
    \scriptsize
    \begin{tabular}{cccccc}
    \toprule
    \multirow{2}{*}{\textbf{Method}} & \multicolumn{2}{c}{\textbf{Communication}} & \multicolumn{2}{c}{\textbf{Memory Cost}} \\
    \cmidrule(lr){2-3} \cmidrule(lr){4-5}
    & \textbf{Cost} & \textbf{Overlap} & \textbf{Model} & \textbf{KV Activations} \\
    \midrule
    
    Ring Attention & $2O(p \times h)L$ & \cmark & $W$ & $F\frac{1}{N}KV$ \\
    
    DeepSpeed-Ulysses & $\frac{4}{N}O(p \times h)L$ & \xmark & $W$ & $F\frac{1}{N}KV$ \\

    Video-Infinity & $2O( Num_{C} \times H' \times W' \times h)L$ & \xmark & $W$ & $(F\frac{1}{N} + Num_{C})KV$\\

    FIFO & $2O( (Num_B+Num_{C}) \times H \times W \times C)$ & \cmark & $W$ & $(Num_B + Num_{C})KV$\\

    \method (Ours) & $2O( (Num_B + Num_{C}/2) \times H' \times W' \times h)$ & \cmark & $\frac{1}{N}W$ & $(Num_B + Num_{C})KV$ \\
    \bottomrule
    \end{tabular}
    \vspace{-0.8em}
}
\end{table}

To compare \method with other parallel methods in terms of communication and memory costs, we qualitatively evaluate it against DeepSpeed-Ulysses~\cite{Ulysses}, Ring Attention~\cite{RingAttention}, Video-Infinity~\cite{infinity}, and FIFO~\cite{FIFO}. Following the approach in previous works~\cite{xDit, Pipefusion}, we conduct a similar comparison for parallelism in video diffusion models, as shown in Table~\ref{tab:parallel_schemes}. 
For \method, the communication cost per device is determined by the input and output of $\epsilon$ through asynchronous P2P communication. Although \method requires synchronous P2P communication during the warm-up and cool-down phases, their overhead is negligible when generating long videos. Furthermore, as detailed in Section~\ref{sec:feature_cache}, we reduce this cost by caching the previous block, resulting in the transmission of only $Num_B + \frac{Num_C}{2}$ frames.
Regarding memory cost, thanks to the advantages of pipeline parallelism, the model cost is distributed across the number of devices, $N$. The memory required for peak $KV$ activation is $(Num_B + Num_C)KV$, which is significantly lower than that of Ring Attention, DeepSpeed-Ulysses and Video-Infinity when generating long videos. This is due to their fixed-length generation nature, which necessitates extending the video sequence at execution time to support long video generation. In contrast, \method and FIFO are infinite-length generation methods that process fixed-length frame blocks at each step and can generate long videos without increasing the number of frames per block. 
In comparison to FIFO, \method shows a substantial memory cost advantage (including both model and KV activations) as the scale of DiT-based video diffusion models increases. Therefore, this quantitative analysis demonstrates the superior performance of \method. Further details of the analysis and calculations are provided in the Appendix~\ref{appendix:efficiency_analysis_in_details}.

%% file: sections/6_experiments.tex
\section{Experiments}\label{sec:experiments}

\subsection{Setups}\label{sec:experiments_setups}
\textbf{Base model.}
In the experiments, the text-to-video model Wan2.1~\cite{Wan} serves as the base model. Wan2.1 is a latest DiT-based video foundation model renowned for its exceptional video generation performance. It is available in two versions: the Wan2.1-1.3B model, which generates 480p videos, and the Wan2.1-14B model, which can generate both 480p and 720p videos. 

\textbf{Metrics evaluation.}
To evaluate parallel efficiency, we compare all methods in terms of generation latency and memory consumption across devices. Memory consumption is measured as the peak memory overhead among all devices used during the diffusion process.
For video performance, we apply VBench metrics~\cite{vbench} directly to long videos~\cite{videomerge}. 
For each method, videos are generated based on the prompts provided by VBench for evaluation. The metrics cover all indicators in the Video Quality category, including subject consistency, background consistency, temporal flickering, motion smoothness, dynamic range, aesthetic quality, and imaging quality.

\textbf{Baslines.}
We benchmark our approach against several existing methods. First, we compare it with Ring Attention~\cite{RingAttention} and DeepSpeed-Ulysses~\cite{Ulysses}, both of which are supported by the official Wan2.1. Additionally, we evaluate it alongside Video-Infinity~\cite{infinity} and FIFO~\cite{FIFO}, two well-established parallel techniques for long video generation. 

\textbf{Implementation details.}
By default, all parameters of the diffusion are kept consistent with the
original inference settings of Wan2.1~\cite{Wan}, with the number of denoising steps set to 50. 
Our experiments are conducted on Nvidia GeForce RTX 4090 (with 24G memory) for Wan2.1-1.3B and Nvidia H20 (with 96G memory and NVLink) for Wan2.1-14B. We utilized the torch.distributed tool package, employing Nvidia’s
NCCL as the backend to facilitate efficient inter-GPU communication. 
We conduct experiments to evaluate the efficiency of both Wan2.1-1.3B (480p) and Wan2.1-14B (720p) in terms of latency and memory usage. For video performance, we compare all methods using the Wan2.1-1.3B (480p) version. For \method, we apply 1 step warmup iteration to make sure the connection between different devices.

\subsection{Main results}\label{sec:experiments_main_results}
\textbf{Efficiency.}
We first evaluate all comparing methods on extremely long videos, followed by scalability analysis. For a fair comparison, we set $Num_C=8$ for \method, Video-Infinity and FIFO, and set $Num_B=8$ for both \method and FIFO.

\begin{wraptable}{r}{0.5\textwidth}
\renewcommand{\arraystretch}{0.5}
\centering
\small
\vspace{-2.em}
\caption{Efficiency evaluation on extreme-length video generation. Experiments are conducted on 8$\times$RTX 4090 GPUs using Wan2.1-1.3B (480p). Results are reported as latency (s) / peak memory usage (GB).}
\resizebox{\linewidth}{!}{
    \begin{tabular}{l>{\centering\arraybackslash}p{2.5cm}>{\centering\arraybackslash}p{2.5cm}}
    \toprule
    \textbf{Methods} & \multicolumn{2}{c}{\textbf{Number of Frames}} \\
    \cmidrule(lr){2-3}
    & \textbf{513} & \textbf{1025} \\
    \midrule
    Ring Attention  & 1328.3 / 17.66 & 3907.5 / 23.15 \\
    Video-Infinity  & 354.1 / 19.13 & 832.5 / 22.58 \\
    FIFO            & 565.3 / 20.59 & 1022.0 / 20.61 \\
    \method         & \textbf{309.3} / \textbf{15.52} & \textbf{596.9} / \textbf{15.55} \\
    \bottomrule
    \end{tabular}
}
\label{tab:extreme_long}
\vspace{-0.5em}
\end{wraptable}
For extremely long videos, as shown in \autoref{tab:extreme_long}, \method achieves great efficiency. Note that DeepSpeed-Ulysses is excluded due to incompatible attention head settings with 8 GPUs.
Compared to static-length generating methods like Ring Attention and Video-Infinity, \method shows clear advantages at 513 frames and further amplifies these advantages at 1025 frames, achieving up to 6.54$\times$ lower latency and 1.48$\times$ lower memory usage in 1025 frames.
This improvement stems from the fact that static-length generating methods require proportionally longer processing sequences as video length increases, leading to higher latency and memory consumption.
In comparison to FIFO—a method for infinite-length video generation—\method still achieves up to a 1.82$\times$ reduction in latency and a 1.32$\times$ reduction in memory usage in 513 frames.
\begin{figure}[t]
    \centering
    \includegraphics[page=1, width=\textwidth]{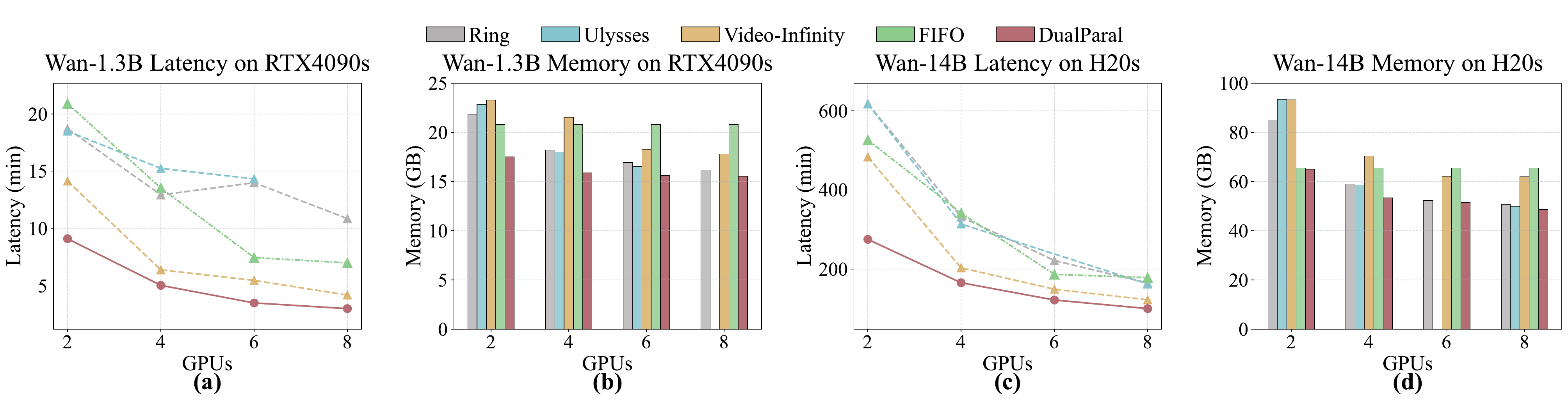}
    \caption{Scalability analysis in terms of latency and memory cost: (a) and (b) show the scalability of Wan2.1-1.3B (480p) across different methods on a 301-frame video, while (c) and (d) present the scalability of Wan2.1-14B (720p) on a 301-frame video.}
    \label{fig:main_results}
    \vspace{-1.em}
\end{figure}

To evaluate scalability, we measure generation latency and memory usage across multiple GPUs using various methods. Experiments are conducted on 301-frame video generation—the maximum length supported by 2 GPUs—and tested on both Wan2.1-1.3B (480p) and Wan2.1-14B (720p) models. As shown in \autoref{fig:main_results}, \method consistently outperforms all methods.
For latency, shown in (a) and (c), \method achieves the lowest generation time across all tested GPU counts. Meanwhile, according to \autoref{eq:bubble}, \method is expected to exhibit even better scalability for longer videos.
For memory usage, shown in (b) and (d), \method maintains the lowest peak memory consumption, with a steadily decreasing trend as the number of devices increases. This efficiency stems from \method’s fixed memory footprint for KV activations and reduced model weight across devices.
In contrast, FIFO shows no scalability in memory usage, posing challenges for large-scale video models. Although Ring Attention, DeepSpeed-Ulysses, and Video-Infinity benefit from reduced memory and latency with more devices, they still face scalability bottlenecks when generating longer videos, as shown in \autoref{tab:extreme_long}.
A more detailed analysis of \autoref{fig:main_results} is provided in Appendix~\ref{appendix:scalability_analysis_in_details}.

\begin{figure}[h]
    \centering
    \includegraphics[page=1, width=\textwidth]{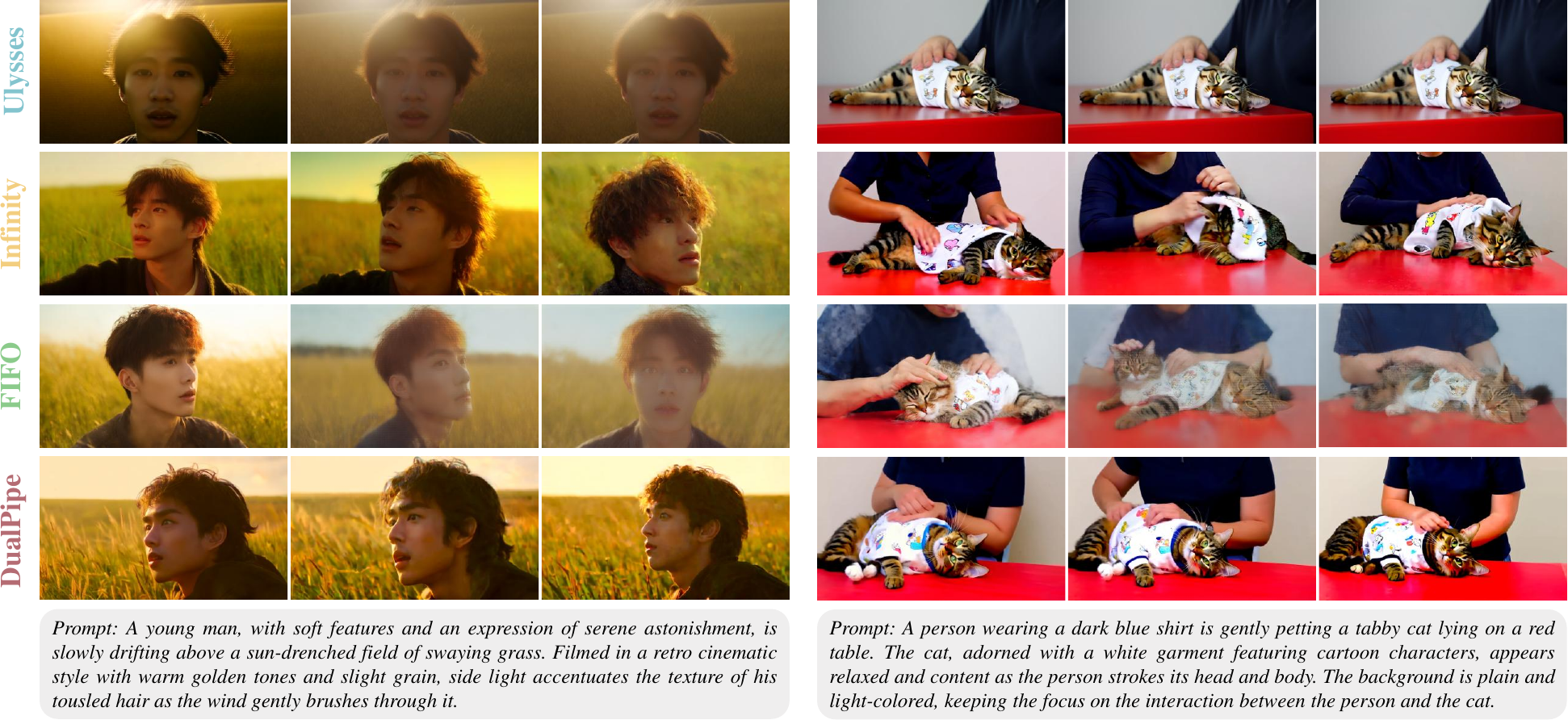}
    \caption{Comparison of 257-frame videos. }
    \label{fig:video}
    \vspace{-0.5em}
\end{figure}

\textbf{Video quality.}
We compare the video quality generated by \method with those produced by DeepSpeed-Ulysses, Video-Infinity, and FIFO on Wan2.1-1.3B (480p) model. \autoref{tab:metrics} presents a quantitative evaluation based on VBench~\cite{vbench}. Additionally, \autoref{fig:video} visualize some frames from videos generated by different methods using the same prompt. 
To ensure optimal video performance, we set $Num_C=24$ with 16 local and 8 global paddings for Video-Infinity, $Num_C=2$ and $Num_B=2$ for FIFO, and $Num_C=8$ and $Num_B=8$ for \method.
\begin{table}[h]
\renewcommand{\arraystretch}{0.8}
\vspace{-1.em}
\caption{The comparison of various video generation methods, as benchmarked by VBench.}
\label{tab:metrics}    
    \resizebox{\linewidth}{!}{
        \centering
        \begin{tabular}{l|c|ccccccc|c}
        \toprule
             \textbf{Method} & \textbf{Number of} &\textbf{Subject}     &  \textbf{Background}   & \textbf{Temporal}      & \textbf{Motion}        & \textbf{Dynamic}  & \textbf{Aesthetic} & \textbf{Imaging}  &\textbf{Overall}\\
              &\textbf{Frames} & \textbf{consistency}& \textbf{consistency}  & \textbf{flickering}    & \textbf{smoothness}    & \textbf{degree} & \textbf{quality} &\textbf{quality}  &\textbf{Score}\\
         \midrule
         \midrule
        \textbf{DeepSpeed-Ulysses} & 129& \textbf{93.43}\%& 92.50\%& 98.88\%& \textbf{98.57}\%& \textbf{62.50}\%& \textbf{62.96}\%& \textbf{64.13}\%& \textbf{81.85}\%\\
        \textbf{Video-Infinity} & 129 & 82.35\%& 88.46\%& 98.41\%& 97.15\%& 59.72\%& 57.69\%& 63.91\%& 78.24\%\\
        \textbf{FIFO} & 129 & 80.46\%& 90.13\%& 95.04\%& 95.30\%& 0\%& 48.28\%& 58.13\%& 75.89\%\\
        \textbf{\method} & 129& 92.92\%& \textbf{95.68}\%& \textbf{99.46}\%& 97.28\%& 59.72\%& 58.86\%& 62.59\%& 80.93\%\\
        \midrule
        \textbf{DeepSpeed-Ulysses} & 257& \textbf{93.45}\%& \textbf{95.07}\%& 98.05\%& \textbf{98.45}\%& 23.61\%& 53.87\%& 55.86\%& 74.05\%\\
        \textbf{Video-Infinity} & 257&86.49\%& 89.41\%& 98.36\%& 97.63\%& \textbf{52.78}\%& \textbf{58.88}\%& \textbf{63.01}\%& 78.08\%\\
        \textbf{FIFO} & 257&71.69\%& 85.41\%& 95.42\%& 95.19\%& 0\%& 48.61\%& 58.18\%& 64.93\%\\
        \textbf{\method} & 257& 89.15\%& 91.80\%& \textbf{99.34}\%& 96.82\%& 50.00\%& 57.35\%& 62.73\%& \textbf{78.17}\%\\
        \bottomrule
        \end{tabular}
    }
    \vspace{-1.em}
\end{table}

\autoref{tab:metrics} reports quantitative results for video generation at 129 and 257 frames. Since both DeepSpeed-Ulysses and Ring Attention operate on full-length sequences without segmentation, only DeepSpeed-Ulysses is selected as a representative case. In the 129-frame setting, DeepSpeed-Ulysses achieves the best performance, as it preserves the full video sequence without splitting, maintaining the original generation quality of Wan2.1. However, its performance drops sharply at 257 frames due to exceeding Wan2.1’s supported video length. In comparison, \method outperforms other distributed methods—including FIFO and Video-Infinity—at 129 frames and achieves the highest overall score at 257 frames. 
For visualization, \autoref{fig:video} presents two example videos. The results align with the quantitative findings in Table~\ref{tab:metrics}. Specifically, directly extending video length—as done by DeepSpeed-Ulysses—causes the Wan2.1 model to produce static scenes lacking motion dynamics. For FIFO, using a single latent per element in queue and denoising under large noise gaps leads to cumulative quality degradation.
In contrast, both Video-Infinity and \method perform well visually. However, Video-Infinity exhibits challenges in maintaining consistency: in the first example, the young man's head orientation shifts erratically; in the second, inconsistencies appear in the depiction of cat and human arm. \method, by comparison, consistently delivers superior temporal coherence across both content and motion. Further video examples are shown in Appendix~\ref{appendix:results_in_long_video}.


\subsection{Ablation}\label{sec:experiments_ablation}
\textbf{Parallel ablation.}

\begin{wraptable}{r}{0.5\textwidth}
\renewcommand{\arraystretch}{0.6}
\centering
\vspace{-1.5em}
\small
\caption{Ablation study on \method. All settings are evaluated on a 129-frame video with 8$\times$4090s.}
\resizebox{\linewidth}{!}{
    \begin{tabular}{lcc>{\centering\arraybackslash}p{2cm}}
    \toprule
    \textbf{Method} & \textbf{Latency (s)} & \textbf{Memory (GB)} & \textbf{Infinite Length} \\
    \midrule
    Queue      & 621.37& 18.81& \cmark \\
    Device Pipeline & 472.25& 21.18& \xmark \\  
    \method$_{w/o\:cache}$ & 182.71& 16.76& \cmark \\
    \method        & \textbf{142.62}& \textbf{15.52}& \cmark \\
    \bottomrule
    \end{tabular}
}
\label{tab:pipeline_compare}
\vspace{-0.5em}
\end{wraptable}
The parallel architecture of \method consists of two main components: queue and device pipeline. By continuously feeding blocks from the queue into the device pipeline, supported by a feature cache mechanism, \method ensures efficient and seamless dual parallelization across devices. To evaluate the individual contributions of each component—queue, device pipeline, and feature cache—we conduct an ablation study focusing on latency, memory usage, and the ability to support infinite-length video generation.
As shown in \autoref{tab:pipeline_compare}, using only the queue with a single GPU results in high latency and memory consumption. In contrast, relying solely on the device pipeline cannot support infinite-length generation and remains inefficient due to underutilization of GPUs when processing a single input. The complete \method architecture, integrating both the queue and device pipeline without cache, successfully addresses these limitations, achieving superior efficiency. With the addition of the feature cache, efficiency is further enhanced, enabling the generation of minute-long videos with ease.

\textbf{Effectiveness of coordinated noise initialization.} 

\begin{wrapfigure}{r}{0.5\textwidth}
\centering
\vspace{-1.8em}
\includegraphics[width=0.48\textwidth]{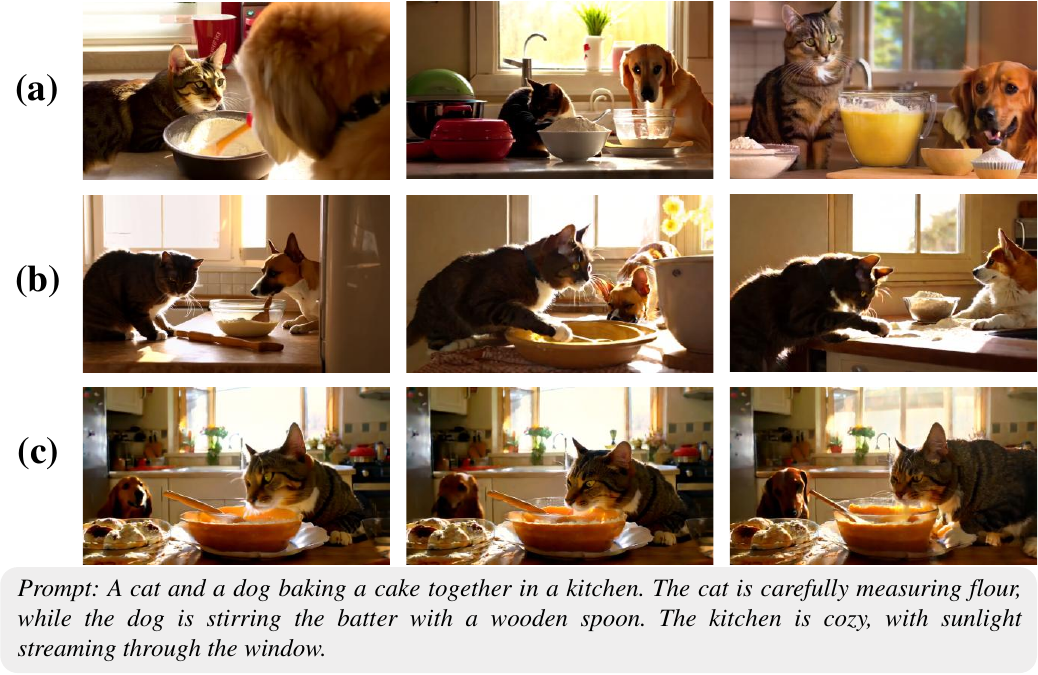}
\caption{Video frames under different conditions: (a) $Num_C=0$ without noise initialization; (b) $Num_C=8$ without noise initialization; (c) $Num_C=8$ with coordinated noise initialization.}
\label{fig:noise_ablation}
\vspace{-1em}
\end{wrapfigure}
Through the two key observations for DiT-based video models discussed in Section~\ref{sec:noise}, we utilize the complete noise space and avoid using the same noise throughout the entire process to bridge the temporal gap between different non-overlapping blocks. In this part, we verify the effectiveness of this approach.
As shown in \autoref{fig:noise_ablation}, without applying noise initialization, \method fails to maintain temporal consistency, as seen in (a) and (b). Although (b) incorporates neighboring blocks to partially mitigate the inconsistency, the effect is limited. In contrast, after introducing noise initialization, as shown in (c), \method achieves significantly improved temporal coherence across frames.

%% file: sections/7_conclusion.tex
\section{Conclusion}\label{sec:conclusion}
We propose \textbf{\method}, a novel distributed inference strategy for DiT-based video diffusion models. 
By implementing a \textit{block-wise denoising} scheme, \method successfully parallelizes both temporal frames and model layers across GPUs, resulting in high efficiency for long video generation.
To further enhance efficiency and video quality, \method reuses KV features via a feature cache strategy for reducing communication and computational redundancy, and applies coordinated noise initialization to ensure global consistency without extra cost. These designs together enable efficient generation of minute-long videos.

%% file: sections/8_appendix.tex
\newpage
\section{Technical Appendices and Supplementary Material}

\subsection{Results in long videos}\label{appendix:results_in_long_video}
More video examples are available on our project page at \url{https://dualparal-project.github.io/dualparal.github.io/}, including minute-long videos, comparisons with other parallel methods, and a diverse gallery. For generating more diverse videos, please refer to our code, available at the repository: \url{https://github.com/DualParal-Project/DualParal}.

\subsection{Related works}\label{appendix:relate_works}

\textbf{DiT-based video diffusion models.}
Recent video diffusion models, including Wan~\cite{Wan}, Hunyuan~\cite{hunyuan}, and OpenSora~\cite{opensora}, have transitioned the architecture from U-Net~\cite{videocrafter1, VideoCrafter2, turbov1, turbov2} to Diffusion Transformers (DiT)~\cite{DiT, ViT_DiT}, a scalable architecture for diffusion models. These models leverage full spatio-temporal attention across all dimensions and perform denoising in the latent space of a pretrained 3D-VAE \cite{hunyuan, 5cogvideox}, enabling more effective extraction of features from video data.
With scaling from 1B~\cite{Wan, opensora} to 14B~\cite{hunyuan, Wan, opensora2} and still growing, DiT-based models have established a strong foundation for video generation.

\textbf{Parallelisms for diffusion models.}
As DiT-based diffusion models scale, their computational and memory demands surpass the capacity of a single GPU, necessitating parallelization. Existing parallel techniques for diffusion models fall into two main categories: sequence parallelism~\cite{Ulysses, RingAttention, infinity, FIFO} and pipeline parallelism~\cite{Pipefusion}.
Sequence parallelism, such as Ring Attention~\cite{RingAttention} and DeepSpeed-Ulysses~\cite{Ulysses}, divides the hidden representations within DiT blocks, enabling parallel attention computation across different attention heads~\cite{Ulysses} or leveraging peer-to-peer (P2P) transmission for Key (K) and Value (V)~\cite{RingAttention}. 
Video-Infinity~\cite{infinity} and FIFO~\cite{FIFO} further extend this idea: Video-Infinity partitions video frames into clips with synchronized context communication across devices, while FIFO introduces a first-in-first-out queue where each element represents a frame at increasing noise levels, enabling infinite-length video generation. Additionally, they still leave room for further efficiency improvements, particularly Ring Attention and DeepSpeed-Ulysses, as they rely on processing the entire video sequence. 
Pipeline parallelism, as in Pipefusion~\cite{Pipefusion}, reduces memory and communication costs by caching image patches across devices for image generation, but struggles with video generation due to the high memory cost in storing patches for all frames~\cite{xDit}.

\textbf{Long video generation.}
The scarcity of long video training data, combined with the high resource cost of generation, results in low-quality and inefficient outputs. 
Although parallelization improves efficiency, it often comes with trade-offs. Methods like Ring Attention~\cite{RingAttention} and DeepSpeed-Ulysses~\cite{Ulysses}, which use full-sequence attention, tend to produce static videos. In contrast, approaches such as FIFO~\cite{FIFO} and Video-Infinity~\cite{infinity}, which partition the input video into non-overlapping blocks, suffer from poor temporal consistency.
Beyond parallelization, noise initialization offers an efficient alternative. FreeNoise~\cite{freenoise} initializes each split block using a subset of the noise space. More recently, VideoMerge~\cite{videomerge} initializes each block with the full noise space but requires latent fusion to ensure coherence between adjacent chunks. 
However, more favorable strategy for noise initialization in DiT-based video diffusion models remains underexplored. Furthermore, how to parallelize noise initialization without compromising generation quality has been largely overlooked.

\subsection{Bubble analysis}\label{appendix:analysis_of_different_denoising_order}
This section provides an in-depth analysis of the bubble ratio introduced in Section~\ref{sec:Quantitative}.
We begin by proving \autoref{eq:bubble} under the reverse denoising order (tail-to-head), assuming $N \leq Block_{num}$.
Next, we examine the bubble ratio in the case where $N > Block_{num}$.
Finally, we analyze the bubble ratio under the sequential denoising order (head-to-tail). 

The proof of \autoref{eq:bubble} consists of two components: the non-bubble size and the bubble size. These correspond to the total execution time and the idle time on each device, respectively.
For the non-bubble size, each block undergoes $T$ denoising steps, and there are $Block_{num}$ total blocks. Hence, the total number of denoising operations is $Block_{num} \times T$. Since every block must traverse all $N$ devices for full denoising, the non-bubble size on each device is also $Block_{num} \times T$. 
Regarding bubble size, idle time in each device will occur in the condition where current number of blocks in queue is smaller than the device number $N$. This condition will occur in warmp-up and cool-down periods when assuming $N\leq Block_{num}$ in \autoref{eq:bubble}. During these periods, as shown in \autoref{fig:Bubble}, the sum of bubble size is equal to $1+2+\dots+(N-1)+ 1+2+\dots+N=N^2-N-1$. Therefore, the whole bubble ratio is equal to \autoref{eq:bubble}. 
\begin{figure}[h]
    \centering
    \includegraphics[page=1, width=\textwidth]{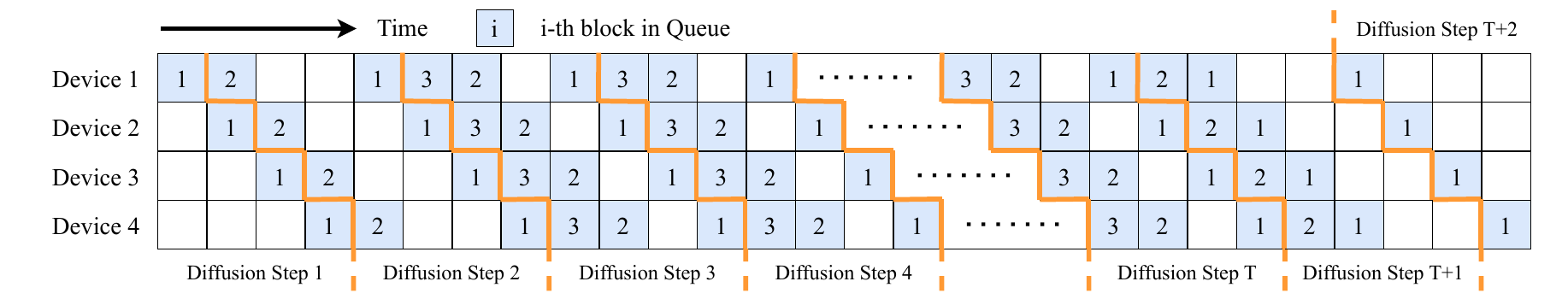}
    \caption{Pipeline schedule of \method with $N=4$, $T=50$, and $Block_{num}=3$. Blocks are denoised  in reverse order, from tail to head in the queue. After diffusion step $T$, the first clean block is popped from the head, and all remaining blocks shift forward by one position, incrementing their indices accordingly.}
    \label{fig:appendix_Bubble_not_full}
\end{figure}

When $N > Block_{num}$, as shown in \autoref{fig:appendix_Bubble_not_full}, bubbles appear in every diffusion step because the number of blocks is insufficient to fully utilize all devices. The bubble ratio in this setting is described as:
\begin{equation}
    Bubble = \frac{Block_{num}*(N-T)+N*(T-2)+1}{Block_{num}*(N-T)+N*(T-2)+1 + T \times Block_{num}}.
\label{eq:bubble_not_full}
\end{equation}
Although the bubble ratio becomes large when $N > Block_{num}$, the condition $N \leq Block_{num}$ is easily satisfied in practice, especially for minute-long video generation. Therefore, we adopt \autoref{eq:bubble} under the assumption $N \leq Block_{num}$ in the main paper to illustrate the bubble ratio.

\begin{figure}[h]
    \centering
    \includegraphics[page=1, width=\textwidth]{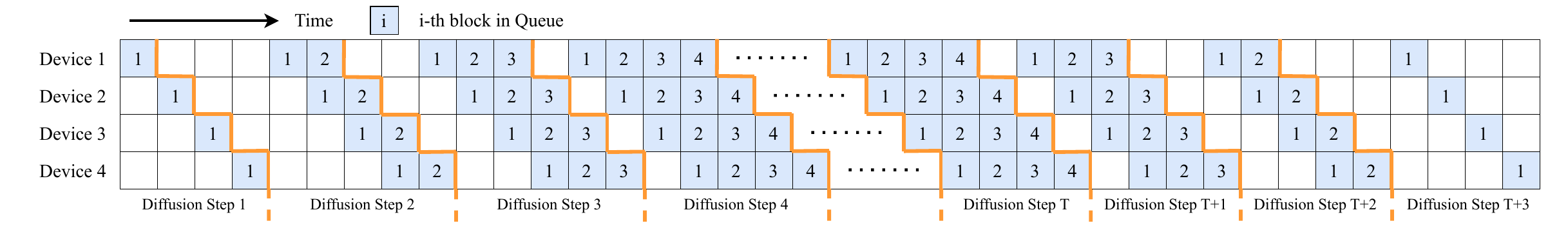}
    \caption{Pipeline schedule of \method with $N=4$, $T=50$, and $Block_{num}=4$. Blocks are denoised sequentially from head to tail in the queue. After diffusion step $T$, the first clean block is popped from the head, and all remaining blocks shift forward by one position, incrementing their indices accordingly.}
    \label{fig:appendix_Bubble_2}
\end{figure}

\method denoises blocks in the queue using a reverse order (from tail to head). To validate the efficiency of this denoising strategy, we further present the bubble ratio under the sequential order (from head to tail) in \autoref{eq:bubble_2} and \autoref{fig:appendix_Bubble_2}. The calculation of the bubble ratio under the sequential denoising order follows a similar approach to that of the reverse order, and we define it as:
\begin{equation}
    Bubble = \frac{Bubble\:Size}{Bubble\:Size + NonBubble\:Size} = \frac{N^2-1}{N^2-1 + T \times Block_{num}}.
\label{eq:bubble_2}
\end{equation}
\autoref{fig:appendix_Bubble_2} will give more details about the pipeline schedule of \method with sequential order. 
By comparing \autoref{eq:bubble} and \autoref{eq:bubble_2}, the reverse order yields a theoretically lower bubble ratio than the sequential order. This theoretical insight is also supported by empirical observations in \autoref{fig:Bubble} and \autoref{fig:appendix_Bubble_2}.
Therefore, we use reverse denoising order for better efficiency.

\subsection{Efficiency analysis in detail}\label{appendix:efficiency_analysis_in_details}
This section provides a more detailed quantitative analysis of the efficiency of the comparison methods in \autoref{tab:parallel_schemes}, since Section~\ref{sec:Quantitative} primarily focuses on the analysis of \method. 

For Ring Attention~\cite{RingAttention}, hidden sequences are split across devices, and the full model is replicated on each device. As a result, each device holds $F\frac{1}{N}KV$ and incurs the full model memory cost $W$. To compute attention, each device must perform P2P communication to gather $F\frac{N-1}{N}KV$ from the other devices. Since each $K$ or $V$ tensor contains $p \times h$ activations across all video frames per DiT block, the total communication cost is $2O\left(\frac{N-1}{N}p \times h\right)L$, which approximates to $2O(p \times h)L$ as $N$ increases. Note that when computing split $KV$ chunks is slower than communication, the computation and memory can be overlapped.

For DeepSpeed-Ulysses~\cite{Ulysses}, hidden sequences are also split across devices, and the full model is replicated on each device, resulting in the same memory cost as Ring Attention. Unlike Ring Attention, Ulysses uses All-to-All communication to transform sequence-wise partitioning into attention-head partitioning, enabling parallel attention computation across heads. This process involves three All-to-All transfers of approximately $\frac{1}{N}QKV$ for computing attention, plus one additional All-to-All transfer to reconstruct the split hidden state before attention. Thus, the total communication cost is $\frac{4}{N}O(p \times h)L$, which cannot be overlapped with computation.

In Video-Infinity~\cite{infinity}, the input video sequence is divided into short clips processed across devices, where each device handles $F \frac{1}{N}$ frames using the whole model. To maintain temporal coherence between adjacent clips, each device collects both global and local context key-value (KV) features covering $Num_C$ frames during the attention operation. Consequently, each device requires $(F\frac{1}{N} + Num_C)KV$ costs and incurs the full model memory footprint $W$. Since Video-Infinity does not process the entire video sequence on each device, we use $1 \times H' \times W'$ to represent the hidden sequence per frame, and $p = F \times H' \times W'$ for the entire sequence. For communication, each device must exchange context features with others at every DiT blocks, resulting in $2O(Num_C \times H'\times W' \times h)L$ overhead, which cannot be overlapped with computation.

For FIFO~\cite{FIFO}, the first-in-first-out queue is split into overlapping blocks, where $Num_B$ frames are denoised with the help of concatenated $Num_C$ frames to maintain temporal coherence. These blocks are processed across devices using the full model. As a result, each device holds $(Num_B + Num_C)KV$ and incurs the full model memory cost $W$. Communication involves distributing each block to all devices and gathering the results on the main device, with a cost of approximately $2O((Num_B + Num_C) \times H \times W \times C)$ to transfer original frame latents, which can overlap with computation. However, the next denoising step must wait until all blocks have completed processing, limiting parallel efficiency. 

For \method, it partitions both video frames and model layers across devices. For memory usage, each device only holds $\frac{1}{N}W$ of the model and $(Num_B + Num_C)KV$ features. Communication overhead primarily arises from P2P transfers of input and output hidden sequences. Thanks to the feature cache technique, only $(Num_B + \frac{Num_C}{2})$ frames need to be transmitted, resulting in a total communication cost of $2O((Num_B + \frac{Num_C}{2}) \times H' \times W' \times h)$. Importantly, except for a few non-overlapping intervals during the warm-up and cool-down phases, most communication and computation can largely overlap throughout the process.

\subsection{Scalability analysis in detail}\label{appendix:scalability_analysis_in_details}

\begin{wrapfigure}{r}{0.5\textwidth}
\centering
\vspace{-1.5em}
\includegraphics[page=1, width=0.48\textwidth]{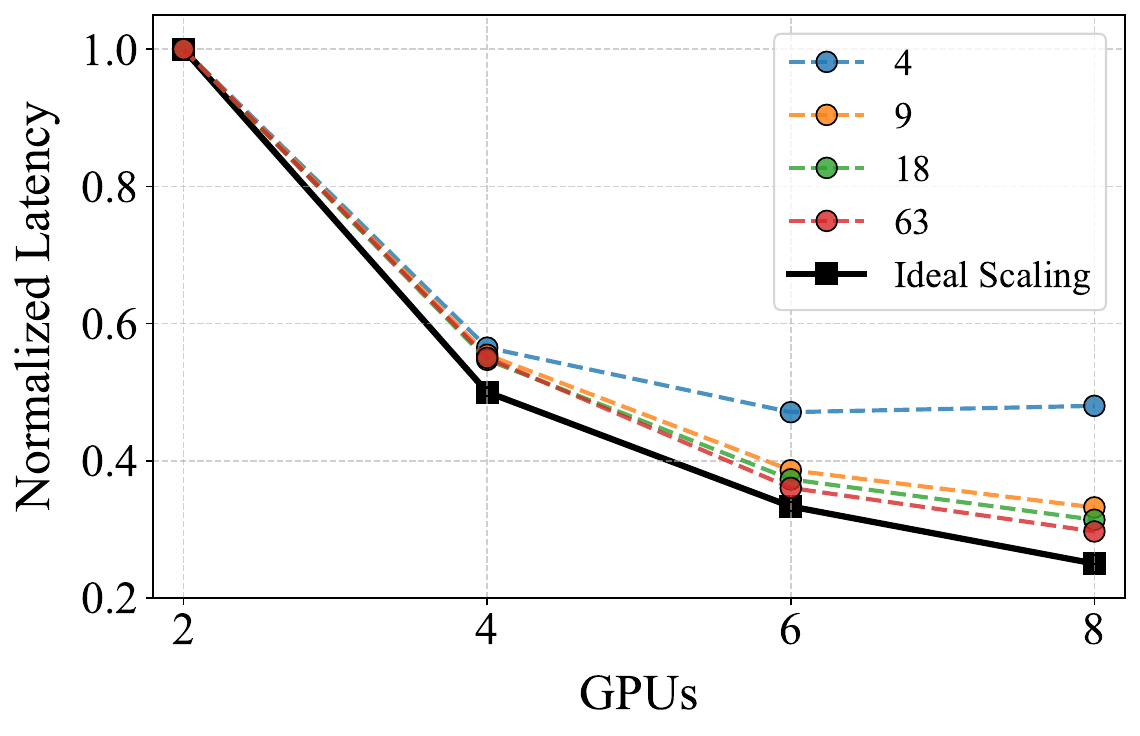}
\vspace{-0.5em}
\caption{The influence of different number of blocks on the scaling ability of \method. Experiments are conducted on Wan2.1-1.3B (480p) using RTX4090s. To better illustrate the scaling behavior, we normalize each line. The black line represents the ideal scaling trend-proportional to the number of GPUs.}
\label{fig:dualpipe_scalability}
\vspace{-1.em}
\end{wrapfigure}
This section provides more detailed analysis of all comparing methods in \autoref{fig:main_results}. 

For \method, as shown in \autoref{fig:main_results}, the full scalability potential in terms of latency is not fully demonstrated due to the limited video length—specifically, when approximately $Block_{num} = 9$ on the 301-frame video used in \autoref{fig:main_results}. This limitation primarily stems from the presence of bubble time on each device, as described in \autoref{eq:bubble}. This additional time includes GPU idling as well as the non-overlapping communication overhead that occurs during the warm-up and cool-down stages.
To further explore the impact of different values of $Block_{num}$ on scalability, we conduct additional experiments. As illustrated in \autoref{fig:dualpipe_scalability}, We evaluate \method with a fixed $Num_C = 8$ across various video lengths, specifically using $Block_{num} = 4$, $9$, $18$, and $63$.
The results show that as $Block_{num}$ increases, the scalability trend increasingly aligns with the ideal scaling—proportional to the number of GPUs. However, when $Block_{num}$ is small, such as $Block_{num} = 4$ (represented by the blue curve), the latency with 8 GPUs surpasses that of using only 6 GPUs. This inefficiency arises because the number of devices $N$ exceeds $Block_{num}$, which prevents \method from effectively implementing seamless workload distribution. Consequently, GPU idling and synchronization overhead are introduced, leading to increased latency.

As shown in \autoref{fig:main_results}, both Ring Attention and DeepSpeed-Ulysses exhibit notable reductions in latency and memory consumption as the number of GPUs increases. In terms of memory usage, their approach by splitting the hidden sequence within DiT blocks reduces the memory required for KV activations as more GPUs are employed. However, despite these improvements, their performance remains inferior to that of \method in both Wan2.1-1.3B and Wan2.1-14B. Moreover, both methods are limited to fixed-length video generation and cannot support infinite-length outputs.
In terms of latency scalability, as shown in \autoref{fig:main_results}(a), Ring Attention experiences increased latency on 6$\times$RTX 4090 GPUs, mainly due to the lack of overlap between communication and computation. Furthermore, DeepSpeed-Ulysses fails to run on 8$\times$RTX 4090s because its number of attention heads is not divisible by 8, rendering multi-head parallelism infeasible. The same issue arises when running on 6$\times$H20 GPUs with Wan2.1-14B.

For Video-Infinity, the method demonstrates suboptimal latency scalability due to its approach of splitting the input video sequence without whole sequence attention. To ensure a fair comparison with FIFO and \method, we set $Num_C = 8$ in our experiments. However, in practical usage, Video-Infinity typically sets $Num_C = 24$, incorporating 16 local paddings and 8 global paddings to improve video quality. This configuration results in a significantly higher latency due to the quadratic scaling with respect to sequence length. In terms of memory usage, the extensive padding used by Video-Infinity also leads to higher memory consumption, exceeding that of both Ring Attention and DeepSpeed-Ulysses. 

For FIFO, an existing method for infinite-length video generation, its efficiency is significantly low. As shown in \autoref{fig:main_results}(a), FIFO demonstrates good scalability when using fewer than 8 GPUs. However, with $Num_B = 8$, the total number of splits in the queue is limited to 7, which is less than 8, thus preventing FIFO from fully utilizing the distributed GPUs. Regarding memory usage, as shown in (b) and (d), FIFO exhibits stability since it fixes $Num_B$ for each chunk and deploys the entire model on every device. However, this approach leads to significant memory issues as the model scales.

\subsection{Limitation}\label{sec:limitation}
The warm-up and cool-down phases in \method are essential for constructing the dual parallelisms but introduce idle time and synchronization overhead. While this overhead is relatively minor when generating long videos, further reducing it could lead to a more optimal and efficient solution.